\pdfoutput=1

\documentclass[11pt]{article}

\usepackage{ACL2023}

\usepackage[notes=true, done=false, later=false, ]{dtrt} 

\usepackage{times}
\usepackage{latexsym}

\usepackage[T1]{fontenc}

\usepackage[utf8]{inputenc}

\usepackage{microtype}

\usepackage{inconsolata}

\usepackage{multicol}
\usepackage{graphicx} 
\usepackage{natbib}  
\usepackage{caption} 
\usepackage{amssymb}
\usepackage[ruled]{algorithm2e}
\usepackage{subfigure}
\usepackage{amsmath}
\usepackage{amsthm}
\usepackage{adjustbox}
\usepackage{multirow}
\usepackage{booktabs}
\usepackage{amsmath}
\usepackage{svg}
\usepackage{arydshln}
\usepackage{cleveref}
\usepackage{wrapfig}
\usepackage{xcolor}
\usepackage{soul}
\crefformat{section}{\textcolor{blue}{\S#2#1#3}} 
\crefformat{subsection}{\textcolor{blue}{\S#2#1#3}}
\crefformat{subsubsection}{\textcolor{blue}{\S#2#1#3}}
\crefformat{equation}{\textcolor{orange}{Eq.~(#1)}}
\usepackage{todonotes}
\usepackage{makecell}
\usepackage{mathtools}
\usepackage{amssymb}
\usepackage{fontawesome5}

\usepackage{thmtools}
\newcommand{\modelname}{\textsc{SubEncoder}\xspace}
\newcommand{\propsegment}{\textsc{PropSegmEnt}\xspace}

\usepackage[most]{tcolorbox}
\usepackage{tcolorbox}
\tcbuselibrary{theorems}

\newtcolorbox{prompt}[1]{
colback=green!5,colframe=green!35!black,fonttitle=\bfseries, title={#1}}

\title{Sub-Sentence Encoder: \\ Contrastive Learning of Propositional Semantic Representations}

\author{Sihao Chen\textsuperscript{$\clubsuit$}\thanks{$\ \ $Work was done during internship at Tencent AI Lab, Bellevue.} \, \textbf{Hongming Zhang}\textsuperscript{$\diamondsuit$} \, Tong Chen\textsuperscript{$\heartsuit$} \, Ben Zhou\textsuperscript{$\clubsuit$} \, Wenhao Yu\textsuperscript{$\diamondsuit$} \\  \textbf{Dian Yu}\textsuperscript{$\diamondsuit$} \quad 
  \textbf{Baolin Peng}\textsuperscript{$\diamondsuit$} \quad \textbf{Hongwei Wang}\textsuperscript{$\diamondsuit$}  \quad \textbf{Dan Roth}\textsuperscript{$\clubsuit$} \quad \textbf{Dong Yu}\textsuperscript{$\diamondsuit$}   \vspace{5pt}
\\ 
  \textsuperscript{$\clubsuit$}University of Pennsylvania \quad \textsuperscript{$\heartsuit$}University of Washington \quad\textsuperscript{$\diamondsuit$}Tencent AI Lab \\
  \small{\texttt{\url{sihaoc@cis.upenn.edu}}}
}

\begin{document}
\maketitle

\begin{abstract}
We introduce \emph{sub-sentence encoder}, a contrastively-learned contextual embedding model for fine-grained semantic representation of text. In contrast to the standard practice with sentence embeddings, where the meaning of an entire sequence of text is encoded into a fixed-length vector, the sub-sentence encoder learns to produce distinct contextual embeddings corresponding to different \textit{atomic propositions}, i.e. atomic units of meaning expressed within a text sequence. The sub-sentence embeddings are contrastively learned to recognize (inferred) semantic equivalence between propositions across different text sequences. Our experiments show the effectiveness of sub-sentence encoders in applications, such as retrieving supporting facts for fine-grained text attribution or recognizing the conditional semantic similarity between texts. In practice, we demonstrate that sub-sentence encoders keep the same level of inference cost and space complexity compared to sentence encoders.

\faGithub \,\,\url{https://github.com/schen149/sub-sentence-encoder}
\end{abstract}

\section{Introduction}
Sentence embeddings are a class of techniques that represent text semantics as dense vector embedding(s) \cite{conneau-etal-2017-supervised, cer-etal-2018-universal, reimers-gurevych-2019-sentence}. Sentence embeddings are widely used in zero-shot or transfer learning settings on information retrieval and text classification tasks \cite{karpukhin-etal-2020-dense, gao-etal-2021-simcse}. 
With sentence embeddings, the common practice is to encode the entire text sequence as a fixed-length vector, where the semantic relation with other text sequences is typically modeled by a similarity function \cite{bromley1993signature}.   

\begin{figure}[t]
    \centering
    \includegraphics[width=\linewidth]{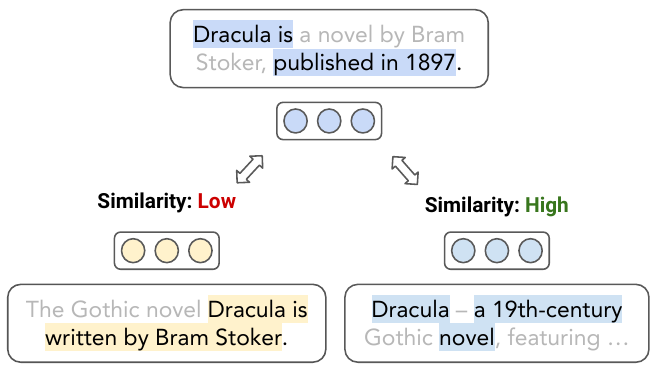}
    
    \caption{Given an atomic proposition in a sentence (represented by a highlighted subset of tokens), the \textit{sub-sentence encoder} produces a contextual embedding for the meaning of the proposition. The cosine similarity between the sub-sentence embeddings captures the (inferred) semantic similarity between the propositions.}
    \label{fig:teaser}
\end{figure}

While sentence embeddings provide unified and compact semantic representations of text, it is difficult to query for the varying granularity of semantics from fixed dimensional sentence embeddings. 
For example, consider the two sentences at the bottom of \autoref{fig:teaser} about the novel \textit{Dracula}. 
While the two sentences as a whole convey different meanings, at the level of \textbf{atomic propositions},
i.e., atomic pieces of meaning conveyed in each sentence, it becomes obvious that the two sentences in part share similar meanings, e.g., both sentences agree on \emph{Dracula is a novel} and \emph{is published in the 19th century}.

Efficiently encoding and indexing text on a more granular level potentially has a profound impact on applications like long-form text evaluation \cite{amplayo2022smart}, attribution \cite{rashkin2023measuring} or factuality estimation \cite{min2023factscore}. With long-form generated text, multiple propositions in the same text might have different truthfulness values. The prerequisite for verifying or attributing such long-form text involves (1) representing the text on a more granular level of atomic propositions and (2) being able to retrieve evidence for different propositions within a text sequence. \cite{chen2023complex, kamoi-etal-2023-wice}.    

\begin{figure*}[t]
    \centering
    \includegraphics[width=\linewidth]{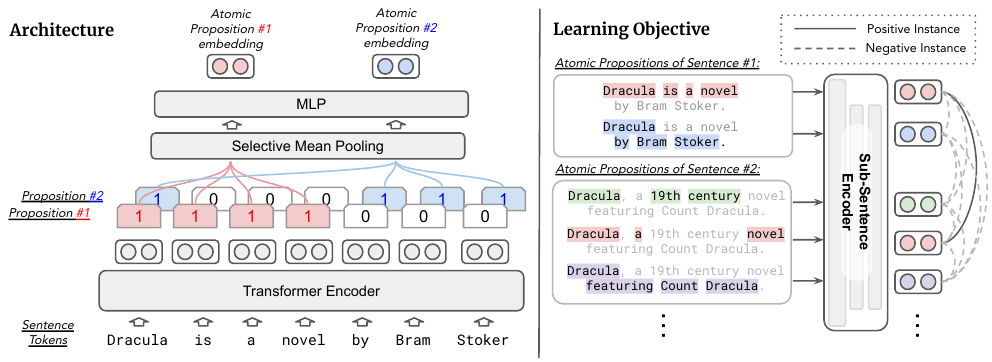}
    \caption{Overview of the \textit{sub-sentence encoder} architecture and learning objective: The model takes a sentence and its propositions (represented as binary token masks) as input and outputs an embedding for each proposition. Given a minibatch of sentences, the model learns to identify pairs of propositions that express the same meaning. All others (including other propositions within the same sentence) are taken as negative examples (\cref{sec:subencoder}). }
    \label{fig:arch}
    \vspace{-5pt}
\end{figure*}
Motivated by such, we introduce \emph{sub-sentence encoder}, a contrastively-learned contextual embedding model for representing sub-sentence-level semantics. As shown in \autoref{fig:arch}, the sub-sentence encoder takes one or more propositions within a text sequence as input. It outputs an embedding that represents the meaning of the proposition. Each proposition takes the format of a binary token mask sequence over the text, which denotes the tokens included in each proposition \cite{chen-etal-2023-propsegment}. 
We train the sub-sentence encoder model to recognize the semantic equivalence between pairs of atomic propositions via in-batch supervised contrastive learning \cite{khosla2020supervised}. We sample and create training examples from a large corpus of unlabeled sentence pair data with proposition extraction and NLI models (\cref{ssec:sampling}).

We evaluate sub-sentence encoders on two types of downstream tasks that involve semantic representation on the sub-sentence level. First, we demonstrate that sub-sentence encoders can be used for fine-grained retrieval, e.g., for text attribution, where a model is expected to retrieve supporting evidence for different parts of a sentence. Second, we show that sub-sentence encoders can be used to infer the conditional semantic similarity between a pair of text \cite{deshpande2023csts}.    

We discuss the design choices and practical challenges in applying sub-sentence encoders in large-scale indexing of a retrieval corpus on the proposition level. As encoding an entire corpus on the proposition level might result in a prohibitively large, we reduce the output dimension of the sub-sentence encoder model during training~\cite{wang2023dimensionality}. We show that this simple yet effective trick results in 12$\times$ to 16$\times$ compression in index size with minimal performance drop. 

The main contributions of the paper are: (1) We propose the sub-sentence encoder, a contextual embedding method for fine-grained text semantics; (2) We introduce an automatic process for creating training data for sub-sentence encoders; (3) We evaluate the utility of sub-sentence encoders in the downstream applications of atomic fact retrieval and conditional semantic textual similarity.  
\section{Preliminaries}
\label{sec:related-work}

\subsection{Motivation: Text Attribution}
Our design of the sub-sentence encoder is largely motivated by the downstream application of text attribution \cite{rashkin2023measuring}, i.e., identify supporting information from known sources to attribute model-generated text. With the widespread adoption of text generation models, evaluating and attributing generated text has become an emerging research topic in need \cite{gao-etal-2023-rarr, gao2023enabling, liu2023evaluating, malaviya2023expertqa}. A key challenge in such tasks lies in the granularity of attributed information, i.e., one piece of generated text usually makes more than one claim, each of which might have different veracity.  For instance, as \autoref{fig:teaser} shows, there could exist multiple claims even within one generated sentence in the form of propositions. Each claim or proposition needs to be contextualized \cite{choi-etal-2021-decontextualization} and individually verified against potentially different information sources \cite{kamoi-etal-2023-wice, min2023factscore}. This process inevitably requires an efficient model representing the semantics of different sentence parts in context, which describes the key design principle for the sub-sentence encoder.

\subsection{Limitations of Sentence Embeddings}
\label{ssec:sent-limitations}
From the perspective of downstream applications such as text attribution, our study addresses the following two shortcomings of current sentence encoder models. 
\paragraph{Granularity.}
Although sentence embeddings usually capture the meaning of the entire text sequence as a fixed-length embedding \cite{morris2023text},
it is difficult in practice to query sentence embeddings for semantic information or structure on a more granular level \cite{rudinger-etal-2017-skip, qin2023nugget, wang2023going}. 
The format would offer limited expressivity when modeling tasks such as document retrieval, especially when the task conceptually involves identifying document parts that respond to the query. 
For such reason, previous studies have found empirical success with phrase retrieval or late-interaction models, which support more granular and expressive representations of the retrieval corpus \cite{seo-etal-2019-real, khattab2020colbert, lee-etal-2021-learning, lee-etal-2021-phrase}. 

\paragraph{Contextualization.}
A typical assumption for sentence encoder models and training/evaluation task setup is that the sentence is encoded independently without context. This becomes a limiting factor in scenarios where similarities and discrepancies between text pairs depend on the context they appear in \cite{chen-etal-2019-seeing, schuster2022stretching, milbauer2023nuisance, milbauer-etal-2023-lait, deshpande2023csts}. 



\section{Sub-Sentence Encoder}
\label{sec:subencoder}
We study a new type of architecture and learning objective -- \textit{sub-sentence encoders}. 
Contrary to sentence encoders, sub-sentence encoders are designed to produce contextual embeddings for each atomic proposition in a sentence. 

\subsection{Architecture}
The sub-sentence encoder architecture is instantiated similarly to transformer-based sentence bi-encoders \cite{reimers-gurevych-2019-sentence}, as shown in \autoref{fig:arch}. The key difference is the sub-sentence encoder takes $k$ sets of binary token masks as extra inputs, which indicate the $k$ propositions of a sentence that it should produce embeddings for. 
The input sentence is first forwarded through a transformer encoder, which can be initialized from any pre-trained encoder model. Then, for each of the $k$ token masks, the token embeddings with mask values of $1$ are mean pooled and forwarded through a projection MLP layer. The model outputs $k$ fixed length embedding corresponding to the $k$ input propositions. 

Note that the $k$ token masks are only applied during pooling, and the encoder still gets full attention to the entire sentence. This allows the proposition embeddings to have the contextual information of the entire input sentence/paragraph, potentially alleviating the need for decontextualizing the propositions \cite{choi-etal-2021-decontextualization}. In addition, since there is no cross-attention between the proposition embeddings, each proposition is encoded independently of others, and its representation is inherently invariant to the input ordering of the propositions.

Compared to sentence encoders, the sub-sentence encoder adds a small amount of parameters with the MLP layer on top. As the sentence can be forwarded only once, the extra inference cost of encoding multiple propositions in a sentence is minimal in practice, as we discuss in \cref{sec:experiments}.

\subsection{Contrastive Learning}
With two propositions from different sentences, the goal is to make them have similar embedding representations if they express similar meanings, and have dissimilar representations otherwise. Within a minibatch of $N$ propositions from $M$ sentences. Let $v_i = \in \mathbb{R}^d$ be the encoded representation of the $i^{th}$ proposition in the batch. Let $I = \{1..N\}$ denote the index of all propositions. 
We formulate the learning objective as minimizing the in-batch supervised contrastive loss $\mathcal{L}$~\cite{khosla2020supervised}: 
\begin{equation*}
    \mathcal{L} = \sum_{i \in I} \frac{-1}{|P(i)|} \sum_{p\in P(i)} \log \frac{\exp(v_i \cdot v_p / \tau)}{\sum_{j\in I \setminus \{i\} }\exp(v_i \cdot v_j / \tau)}
\end{equation*}

\noindent where $P(i)$ is the set of indices of all positive propositions to the $i^{th}$ proposition within the minibatch, and $|P(i)|$ denotes its cardinality. $\tau$ controls the softmax temperature. The learning objective encourages the model to produce embeddings with higher cosine similarity with positive pairs of propositions, while all other propositions in the same batch are considered negatives. Note that if all propositions from the same sentence are packaged in the same minibatch, under the assumption that they cannot be positive examples of each other, the learning objective would inherently encourage the model to assign different representations for different parts of a sentence. 

The supervised contrastive loss is a generalized form of other commonly used loss functions for bi- or dual-encoder training, e.g., N-pairs loss \cite{sohn2016improved} or in-batch softmax \cite{karpukhin-etal-2020-dense}. We opt for this formulation mostly due to its ability to generalize to an arbitrary number of positive examples in the same batch. In our case, this is important, as each proposition may have zero or more positive instances in the same minibatch.   

\subsection{Sampling Proposition Pairs for Training}
\label{ssec:sampling}
Here, we describe how we automatically sample positive proposition pairs from a collection of unlabeled sentence pairs as training data for the sub-sentence encoder. 
We start from a collection of 2.5M sentence pairs from topically related news articles~\cite{zhou-etal-2022-learning-decompose}. This data is collected from RealNews \cite{Zellers2019DefendingAN} to find parallel sentence pairs that generally describe the same event with slightly different angles and focuses. These instances serve as great starting points for our need since we want to find proposition pairs with both similarities and differences. 

\paragraph{Step 1: Segment Sentences $\Rightarrow$ Propositions. } Given an unlabeled sentence pair, we first parse each sentence into propositions in natural language forms. 
First, we prompt \texttt{GPT-3.5-turbo} to generate propositions for $1\%$ of all sentence pairs as the seed set of training data. We find that \texttt{GPT-3.5-turbo} with few shot in-context demonstrations gives reasonable performance on the task, which echos with the observations from \citet{min2023factscore, kamoi-etal-2023-wice}. 

Next, we finetune T5-large \cite{2020t5} on the seed training set and use the model to generate propositions for the rest of the dataset.  
We include more details about the prompt and training process in \autoref{app:segment}.
\paragraph{Step 2: Identify Positive Pairs with NLI models. } 
Given the two sets of propositions (in natural language form) in each sentence pair, we infer and label the positive proposition pairs with an off-the-shelf NLI model \cite{nie-etal-2020-adversarial} \footnote{\url{https://huggingface.co/ynie/roberta-large-snli_mnli_fever_anli_R1_R2_R3-nli}}. We forward each pair of propositions across two sentences through the NLI model two times, with flipped orders between hypothesis and premise. We label a proposition pair positive if the NLI model classifies their relation as entailment in both directions. We only keep sentence pairs with at least one pair of positive propositions. This leaves us with $240k$ sentence pairs, with $3.32$ propositions per sentence and $1.21$ positive propositions on average.

\paragraph{Step 3: Convert Propositions $\Rightarrow$ Token Masks. } 
We convert the propositions in natural language form to the token mask format used for sub-sentence encoder input by aligning the tokens in each proposition to the sentence. We use NLTK \cite{bird2009natural} to lemmatize each token in a proposition and its sentence and construct an affinity matrix between the two, where tokens with identical lemmas are assigned a similarity score of 1. To break ties between multiple token matches, we apply a 2D-convolution filter on the affinity matrix, which adds a small score offset for other token matches in a context window of three tokens. We find the optimal matches between the proposition and sentence with max bipartite matching on the affinity matrix with the Hungarian algorithm \cite{kuhn1955hungarian}. We include a more detailed process description in \autoref{app:segment}.

\section{Experimental Setup}
\label{sec:experiments}
\subsection{Model Configurations}
\label{ssec:model}

\begin{table*}[!t]\centering \small
\begin{tabular}{@{}c|c|cccc@{}}
\toprule
System & Param. Size & \textsc{Precision@1} & \textsc{Recall@5}  & \textsc{Recall@10} & \textsc{Recall@20}\\ 
\midrule
MiniLM-L6-v2 & 23M & 18.36 & 37.28 & 44.87 & 51.62 \\
DistilRoberta & 82M & 16.59 & 33.65 & 40.82 & 46.79 \\
SimCSE (unsupervised) & 110M & 8.90 & 45.13 & 69.47 & 84.29 \\
SimCSE (supervised) & 110M & 16.53 & 57.83 & 77.28 & \textbf{87.89} \\
GTR$_{base}$ & 110M & 21.90 & 52.50 & 65.54 & 75.69 \\
ST5$_{base}$ & 110M & 26.16 & 57.65 & 69.00 & 78.58 \\
\midrule
\modelname (SimCSE) & 110M \textcolor{blue}{(+0.5M)} & \textbf{41.64} & 71.48 & 78.22 & 83.34 \\
\modelname (ST5$_{base}$) & 110M \textcolor{blue}{(+0.5M)} & 40.97 & 72.15  & 79.30 & 84.33 \\
\modelname (GTR$_{base}$) & 110M \textcolor{blue}{(+0.5M)} & 40.77 & \textbf{72.90} & \textbf{80.45} & 85.81 \\

\bottomrule
\end{tabular}
\caption{Zero-shot evaluation results on the \emph{Atomic Fact Retrieval} task in \propsegment \citep{chen-etal-2023-propsegment}. }
\label{tab:propsegment}
\end{table*}

\begin{figure}[t]
    \centering
    \includegraphics[width=0.8\linewidth]{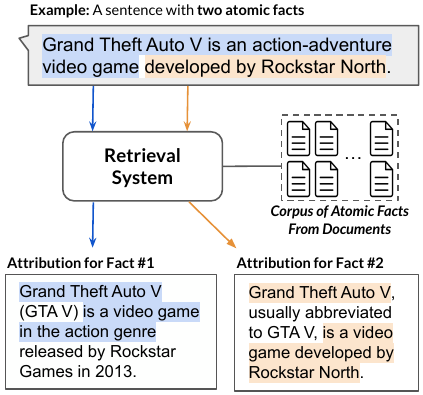}
    \caption{In the \textit{Atomic Fact Retrieval} task \cite{chen-etal-2023-propsegment}, given an atomic proposition in the query, a system is expected to retrieve the set of supporting atomic propositions from the corpus. The dataset features $8.8k$ query propositions and $45k$ candidate evidence propositions from $1.5k$ documents.}
    \vspace{-5pt}
    \label{fig:retrieval}
\end{figure}

We initialize the transformer encoder layers with pre-trained weights from three types of sentence encoders: SimCSE~\cite{gao-etal-2021-simcse}, Sentence-T5~\cite{ni-etal-2022-sentence}, and GTR \cite{ni-etal-2022-large}. With Sentence-T5 and GTR, we experiment with the base, large, and xl-sized variants of the models. For the MLP layer, we keep the output dimension the same as the transformer encoder. We discuss the impact of varying output dimensions in~\cref{ssec:compress}. 

We finetune the sub-sentence encoder with different variants of backbone sentence encoders on the $240k$ sentence pairs with at least one pair of positive propositions. We denote the resulting model as \modelname. We include the details for our distributed training setup and hyperparameters in \autoref{app:hyperparam}. 

\subsection{Evaluation}
To assess the utility of the sub-sentence encoders, we evaluate our model on two types of downstream tasks in zero-shot settings. 
\label{ssec:evaluation}

\subsubsection{Atomic Fact Retrieval for Fine-Grained Text Attribution}
We first evaluate the sub-sentence encoders in retrieving fine-grained attributions \cite{rashkin2023measuring} for text. We conduct the evaluations with the \propsegment dataset \cite{chen-etal-2023-propsegment}. An overview of the task setup is shown in \autoref{fig:retrieval}. Given an atomic proposition in the sentence, a system is expected to identify and retrieve supporting evidence from a corpus of  $\tilde~45k$ human-labeled atomic propositions from $1.5k$ News or Wikipedia documents in total. The task setup emulates the setting where each part of a sentence might have different veracity, and so each atomic proposition in a sentence might be attributed to different supporting evidence from different source documents. On average, each query proposition has $1.13$ ground truth supporting propositions. 

\paragraph{Metrics.} Given a system's output rankings of the $45k$ candidate evidence propositions with respect to a query proposition, we measure the precision@$1$ plus recall@$\{5, 10, 20\}$ of the ranking against the human-annotated ground truth set of evidence propositions.    

\paragraph{Baselines.} We compare pre-trained sentence encoders as baselines. We first evaluate variants of unsupervised and supervised SimCSE, Sentence-T5, and GTR on similar model parameter sizes. In addition, we compare two popular compact models, i.e., \texttt{all-MiniLM-L6-v2} and \texttt{all-distilroberta-v1} from \texttt{sentence-transformers} \cite{reimers-gurevych-2019-sentence}. We discuss the setup for sentence encoders for the tasks in \autoref{app:evaluation}.

\paragraph{Results.}
\autoref{tab:propsegment} summarizes our evaluation results. 
We observe that the \modelname with different backbone encoders generally improve over their sentence encoder counterparts. 
We see the most visible improvements of \modelname in terms of Precision@1 and Recall@5, while the performance gap becomes smaller in terms of Recall@10 and 20. 
This suggests that sub-sentence contrastive learning gives the model better capabilities at recognizing the nuanced semantic differences between propositions appearing in the same context.  

Across different variants of \modelname with different backbone sentence encoders, we observe similar performance levels overall, with the GTR$_{base}$ variant having a slight edge. In \autoref{tab:propsegment}, we mostly compare models with the same backbone encoder size and configurations. Our \modelname only introduces 0.5\% extra parameters with the MLP layer on top.   
We include a more comprehensive analysis of model size, efficiency, and performance trade-off in \cref{sec:discussion}.   

\begin{table*}[t]\centering\scriptsize
\begin{tabular}{p{0.15\linewidth}p{0.25\linewidth}p{0.25\linewidth}p{0.12\linewidth}cc}
\toprule
     Type & Sentence 1 & Sentence 2 & Condition & Label & Pred.  \\
\midrule
\textcolor{teal}{Correct.} & A group of people go \hl{sledding} on a snowy hill, and a dog chases one as he \hl{slides}. & A person, dressed in black, \hl{skipping} down a snow covered road and \hl{playing} with a black dog. & The physical activity. & 4 & \textcolor{teal}{3.44} \\
\midrule
\textcolor{red}{Mistake}: fails to find a good set of words. & A man being thrown into the air while being trampled by a bull. & The cowboy \hl{holds on} to the bull who is \hl{desperately trying to throw him off}. & The person's elevation. & 4 & \textcolor{red}{1} \\
\midrule
\textcolor{red}{Mistake}: correct set of words; failed inference. & A \hl{man wearing a white tank top and a white hard hat} is holding two pieces of pipe at a \hl{construction site}. & A \hl{construction worker} in a lime-green safety vest and orange hard hat is looking closely at something held in his hands. & The occupation of the man. & 5 & \textcolor{red}{2.47} \\

\bottomrule
\end{tabular}
\caption{Example outputs and typical mistakes of \modelname on C-STS. The set of words identified by \texttt{gpt-3.5} is highlighted \hl{yellow}. For display purposes here, the model predicted cosine similarity is normalized to match the human labels' scale of 1 - 5, where 1 $=$ Least similar, and 5 $=$ Most similar}
\label{tab:csts_examples}
\end{table*}

\begin{table}[t]\centering\small
\begin{tabular}{clc}
\toprule
Model & Setting & Spearman $r\uparrow$ \\
\midrule
Roberta$_{base}$ & \textit{0-shot (No. Cond.)} & -0.43$^*$ \\
SimCSE$_{base}$ & \textit{0-shot (No. Cond.)} & 1.66$^*$ \\
\cmidrule{2-3}
\multirow{ 2}{*}{\texttt{FlanT5$_{large}$}} & \textit{0-shot} & -3.0$^*$ \\
& \textit{2-shot} & 11.7$^*$  \\
\cmidrule{2-3}
\multirow{ 2}{*}{\texttt{GPT-3.5}} & \textit{0-shot} & 14.1 \\
& \textit{2-shot} & 15.4 \\
\cmidrule{2-3}
\multirow{ 2}{*}{\texttt{GPT-4}} & \textit{0-shot} & 36.9 \\
& \textit{2-shot} & \textbf{40.7}  \\
\midrule

\multirow{3}{*}{\makecell{\texttt{GPT-3.5} \\ + \modelname \\ (0-shot)}} & (SimCSE$_{base}$) & 27.5 \\
& (GTR$_{base}$) & 31.9 \\
& (ST5$_{base}$) & \textbf{33.0} \\

\midrule
\multirow{3}{*}{\makecell{\texttt{GPT-4} \\ + \modelname \\ (0-shot)}} & (SimCSE$_{base}$) & 34.5 \\
& (GTR$_{base}$) & 36.9 \\
& (ST5$_{base}$) & 37.2 \\
\bottomrule
\end{tabular}
\caption{Spearman correlation coefficient ($\times100$) of model predictions evaluated in zero- or few-shot settings on the \textit{Conditional Semantic Textual Similarity} (C-STS) task. * denotes results from \citet{deshpande2023csts}.}
\vspace{-5pt}
\label{tab:csts}
\end{table}

\subsubsection{Conditional Semantic Text Similarity}
\label{sssec:csts}
To assess \modelname's ability to produce contextual representations for fine-grained sub-sentence level semantics of text, we conduct experiments on the Condition Semantic Text Similarity (C-STS) task \cite{deshpande2023csts}. Compared to STS \cite{agirre-etal-2012-semeval}, C-STS introduces a condition notion of similarity between text pairs, where an additional natural language condition is provided along with the text pair as input. A system is expected to output a similarity score between the pair from the perspective of the given condition. \autoref{tab:csts_examples} shows some examples of the task.

\paragraph{Method.} Given the condition, we first prompt an LLM to identify a set of words in each sentence that best corresponds to the condition. Here, the LLM only sees one sentence at a time, so the condition words in each sentence are identified independently.  We use the sub-sentence encoder to encode the set of words in the context of each sentence as the conditional representation. We take the cosine similarity between two encoded sets of words from the text pair as their conditional similarity.

\paragraph{Metrics.} We compare the Spearman correlation coefficient between predicted similarity from a system against human ratings.  

\paragraph{Baselines.} We compare to a list of zero- and few-shot baselines provided by \citep{deshpande2023csts}. This includes two bi-encoder models, Roberta$_{base}$ and SimCSE$_{base}$ that do not make use of the condition as input, as well as zero- and few-shot prompting results with FlanT5$_{large}$, \texttt{GPT-3.5-turbo} and \texttt{GPT-4}, where each LLM provided detailed instructions of the task, and is prompted to generate a similarity score from 1 to 5 given the text pair and condition with/without in-context demonstrations.

\paragraph{Results.}
\autoref{tab:csts} shows the evaluation results. By having \modelname comparing the contextual similarity between the set of words selected by \texttt{gpt-3.5-turbo}, we see an improvement in zero-shot setting from Spearman's $r = 14.1 \rightarrow 33.0$, compared to directly prompt LLM to output the similarity. However, the performance gap when using \texttt{gpt-4} becomes much smaller ($r = 36.9 \rightarrow 37.2$). This is reasonable considering the fact that direct \texttt{gpt-4} prompting demonstrates on-par performance with supervised systems on C-STS, as reported in \citet{deshpande2023csts}.
We show examples of typical mistakes made by our model in \autoref{tab:csts_examples}. We observe that our method typically fails when (1) the LLM fails to identify a good set of condition words, or when no such corresponding words explicitly exist in the sentence, or (2) \modelname fails to correctly infer the similarity between the two sets of condition words. For instance, with the third example in \autoref{tab:csts_examples}, the inference is particularly challenging for the model, considering the relation between two sentences is modeled only via a cosine similarity with no learned parameters.




\section{Analysis and Discussions}
\label{sec:discussion}
\begin{figure}[t]
    \centering
    \includegraphics[width=\linewidth]{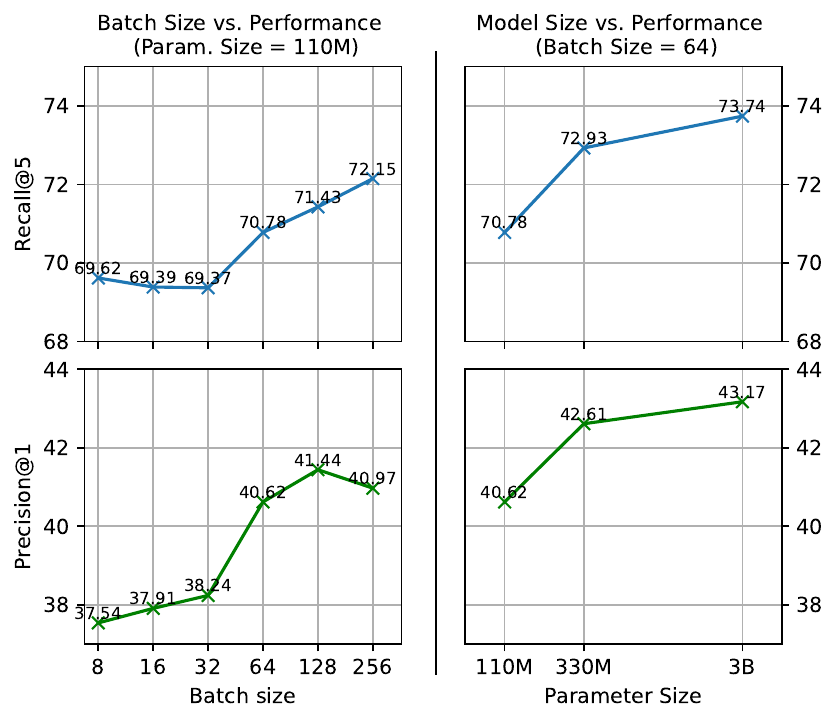}
    \caption{The effect of varying batch size and model parameter size on the atomic fact retrieval performance, tested with the Sentence-T5 variant of \modelname. }
    \label{fig:scaling}
    \vspace{-5pt}
\end{figure}
\subsection{Scaling \modelname}
In \autoref{fig:scaling}, we show an analysis of the effect of scaling model sizes and batch sizes during training. For our analysis, we use the Sentence-T5 variant of \modelname, and evaluate the performance on the atomic fact retrieval task. 

\paragraph{Scaling Batch Size.} As our contrastive learning objective leverages in-batch negative sampling, scaling up the batch sizes during training could bring performance gains. To illustrate this, we initialize \modelname with Sentence-T5 base encoder parameters and finetune with a varying batch size of $\{8, 16, 32, 64, 128, 256\}$. 
We observe that increasing the batch size generally increases performance, which suggests that batch size scaling could yield better model generalizability. 
We observe a significant performance gain when increasing batch size from $32 \rightarrow 64$ while seeing diminishing gains with further increase. This echoes the empirical findings with in-batch contrastive learning in general \cite{khosla2020supervised}. The phenomena can be attributed to the model-predicted labels in our training dataset, which can be noisy. 

\paragraph{Scaling Model Size.} We initialize the encoder with different sizes of Sentence-T5 from 110M to 3B parameters and finetune with a fixed batch size of 64. We observe that starting from a larger pre-trained encoder brings better performance. We see a bigger gain when increasing the model size from 110M to 330M, while the gain becomes smaller when we increase from 330M to 3B.

\subsection{Using \modelname for Sentence or Document Retrieval}
In \cref{sec:experiments}, we compare \modelname's performance on the atomic fact retrieval task against the baseline sentence encoders. In reality, a more likely application scenario is when the system is expected to retrieve supporting evidence on the sentence or document level. To evaluate this, we cast the atomic fact retrieval task as a sentence or document retrieval task, where given a query proposition, a system is expected to retrieve the set of sentences or documents that contain the target proposition(s).  

From the intuition that finer-grained retrieval, e.g., with propositions, entails the more coarse sentence- or document-level retrieval, we follow \citet{lee-etal-2021-phrase} and use a simple strategy with \modelname for sentence- and document-level retrieval. Given each query, we retrieve a slightly larger number of propositions. From the set of sentences and documents where the propositions belong, we use the highest score among the set of propositions as the score for each sentence or document. The top $k$ unique sentences or documents are then returned as results.

\autoref{tab:multi-grain} shows the the evaluation result. Compared to GTR$_{base}$ and Sentence-T5$_{base}$, which are trained for document-level and sentence-level retrieval, respectively, we observe a similar level of performance with retrieving by propositions with \modelname. Overall, we see lower top-1 accuracy compared to the baselines. This is possibly due to the more complex nature of the proposition retrieval task. However, we generally see an improvement in terms of recall @ 5.  The findings indicate the potential of using \modelname for multi-vector retrieval across different granularities.  

\begin{table}[t]\centering\small
\begin{tabular}{c|cc|cc}
\toprule
\multirow{2}{*}{Model} & \multicolumn{2}{c|}{Sentence-Level} & \multicolumn{2}{c}{Document-Level} \\
& P@1 & R@5 & P@1 & R@5 \\
\midrule

GTR$_{base}$ & 49.35 & 77.01 & \textbf{51.93} & 81.97 \\
Sentence-T5$_{base}$ & \textbf{50.59} & 79.37 & 45.27 & 77.10 \\
\midrule
\makecell{\scriptsize \modelname (GTR)} & 42.94 & \textbf{82.27} & 45.04 & \textbf{90.13} \\
\makecell{\scriptsize \modelname (ST5)} & 43.49 & 81.44 & 45.93 & 89.19 \\

\bottomrule
\end{tabular}
\caption{Sentence and document retrieval performance of the atomic fact retrieval task. We evaluate GTR$_{base}$ and Sentence-T5$_{base}$ variants of \modelname.}
\vspace{-5pt}
\label{tab:multi-grain}
\end{table}

\subsection{Robustness to Input Formats/Boundaries}
Although \modelname is fine-tuned with data formatted as propositions specifically, we observe from the C-STS evaluations that the model generalizes to not necessarily proposition-shaped inputs, as shown in \autoref{tab:csts_examples}. In downstream applications, we would expect the model to generalize to input token masks with imperfect boundaries, e.g., propositions generated by a model instead of labeled by humans. 
Alongside the C-STS evaluation results, which indirectly support our hypothesis, we conduct a simple evaluation with the atomic fact retrieval task. Instead of human-annotated queries, we use queries generated by \texttt{gpt-3.5-turbo}. The evaluation results of the proposition segmentation performance of \texttt{gpt-3.5-turbo} and the distilled T5-Large model can be found in \autoref{app:segment}. We observe that, with a fuzzing token-level Jaccard-similarity-based metric, most propositions extracted by the model can be aligned with human-labeled propositions. When we test the atomic fact retrieval performance on the set of model-generated propositions that can be fuzzy-matched with the human-annotated ones, we only see a small drop in performance, e.g., with the GTR-base variant of \modelname, precision@1 drops from 41.21. $\rightarrow$ 39.56, recall@5 drops from 73.14 $\rightarrow$ 72.23. We hypothesize that the robustness of proposition boundaries partly comes from how we train the model. As the labels between proposition pairs are generated independent of the model-generated fuzzy proposition boundaries, the model potentially learns to adapt to imperfect proposition boundaries, similar to the intuition behind unsupervised SimCSE training \cite{gao-etal-2021-simcse}. 
\begin{table}[t]\centering\small
\begin{tabular}{c|cll}
\toprule
Model & Dim. & Precision@1 & Recall@5 \\
\midrule

\multirow{2}{*}{\makecell{\modelname \\ \scriptsize (ST5-Large)}} & 1024 & 42.61 & 72.93 \\
& 64 & 42.10 \textcolor{red}{(-0.51)} & 70.17 \textcolor{red}{(-2.76)}\\

\midrule
\multirow{2}{*}{\makecell{\modelname \\ \scriptsize  (ST5-Base)}} & 768 & 40.97 & 72.15 \\
& 64 & 40.45 \textcolor{red}{(-0.52)} & 71.62 \textcolor{red}{(-0.53)} \\
\bottomrule
\end{tabular}
\caption{The performance difference on the atomic fact retrieval task with vs. without reducing the output dimensionality of \modelname.}
\vspace{-5pt}
\label{tab:compress}
\end{table}

\begin{table}[t]\centering\small
\begin{tabular}{cccc}
\toprule
Index & Num. Entries & Dim. & Index Size \\
\midrule

Propositions & 270M & 64 & 62GB \\

\texttt{dpr-100w} & 21M & 768 & 61GB \\

\bottomrule
\end{tabular}
\caption{The resulting index size with proposition-level indexing with compressed dimension, compared to a DPR index of 100-word blocks \cite{karpukhin-etal-2020-dense}. }
\vspace{-5pt}
\label{tab:index-size}
\end{table}

\subsection{Offline Indexing and Compression}
\label{ssec:compress}
With the promising performance gain from \modelname in the atomic fact retrieval task, we discuss and assess the possibility of applying \modelname for fine-grained retrieval on larger-scale corpora, which involves offline indexing and caching the encoded corpora. In our case, the indexing happens on the level of atomic propositions, where we need to store one embedding for every atomic proposition in the corpora. Compared to document-level indexing, indexing on the proposition level would result in a prohibitively large index size. In previous works \cite{lee-etal-2021-phrase}, this is commonly addressed with techniques such as product quantization \cite{jegou2010product} to compress index size or approximate nearest neighbor search \cite{malkov2018efficient} for faster inference. 

Orthogonal to the two techniques above, we study a simpler yet effective compression strategy by reducing the output dimension of \modelname. In the context of sentence encoders, \citet{wang2023dimensionality} discover that reducing the output dimensionality during training generally incurs minimal downstream performance loss. Following this idea, we finetune the Sentence T5 base and large variants of \modelname with a bottlenecked output dimension of 64 instead of the original output dimensions of 1024 and 768, respectively. \autoref{tab:compress} shows the performance comparison when evaluated on the atomic fact retrieval task. Overall, we observe a very small performance drop while gaining 12$\times$ to 16$\times$ reduction in output embedding size.

To demonstrate the implication of this in practice, we use the Sentence T5 large variant of \modelname to encode and index an English Wikipedia dump from 2021/10/13, as used by \citet{bohnet2022attributed}. We segment all sentences in Wikipedia into propositions with the T5-large model (\cref{ssec:sampling}). This results in $\tilde~270$M propositions from 5.3M Wikipedia pages. \autoref{tab:index-size} shows the resulting index. The resulting size of 62GB is close to a prebuilt (uncompressed) dense passage retrieval (DPR) index on the level of 100-word blocks \cite{karpukhin-etal-2020-dense}. We see that decreasing the output dimension of the embeddings helps in reducing the cached index size. It is worth noting that compared to the document-level index, we still expect the query speed of the index to increase slightly due to the increase in the number of entries. However, in practice, we observe a reasonable overall time and space complexity involved in offline indexing and online similarity querying on the proposition-level.

\section{Conclusion}
\label{sec:conclusion}
We introduce sub-sentence encoders, a contrastive learning framework for learning contextual embeddings for semantic units on the sub-sentence level. Beyond the use cases covered in the paper, the sub-sentence encoder architecture could potentially serve as the backbone for any cross-document information linking tasks in context, and the learning objectives could potentially apply to a broader range of tasks with various granularity of information, e.g., linking sentences or spans within different documents \cite{ma-etal-2023-chain}. We hope that the findings in this paper will facilitate further exploration along these directions. 
    
\section*{Limitations}
This work mostly serves as exploratory work to validate the idea behind sub-sentence encoder architecture and learning objectives. 
We acknowledge the limited scale of our experiments, specifically in terms of the \textit{languages supported by the model}.
In our experiments, we explore the idea of \emph{sub-sentence encoder} with English text only. However, the techniques described in the paper for sampling training data and training the \emph{sub-sentence encoder} can be applied to other languages as well. We leave exploration on multilingual \emph{sub-sentence encoder} for future work. 
\section*{Acknowledgements}
The authors would like to thank Alex Fabrikant, Jianmo Ni, and Tal Schuster for the discussions leading to the development of this idea. The authors thank Xinran Zhao, Kaixin Ma, Vivek Gupta, and Xiaodong Yu for valuable feedback on the project and the paper presentation.
\bibliography{anthology,custom}
\bibliographystyle{acl_natbib}

\newpage
\appendix

\begin{table*}[t]\centering\small
\begin{tabular}{ccccccc}
    \toprule
     \multirow{ 2}{*}{Model} & \multicolumn{3}{c}{Jaccard $\theta=0.8$} & \multicolumn{3}{c}{Jaccard $\theta=0.5$} \\
       & Precision & Recall & F1 & Precision & Recall & F1 \\
\midrule
\multicolumn{7}{c}{\textit{Systems used in this paper}} \\
\midrule
\texttt{GPT-3.5-turbo} & 35.79 & 31.65 & 33.60 & 71.52 & 63.87 & 67.48\\
\textsc{T5}-Large (w/ GPT3.5 training data) & 35.91 & 31.70 & 33.68 & 70.27 & 63.39 & 66.65 \\
\midrule
\multicolumn{7}{c}{\textit{Systems fine-tuned on 
 \propsegment \cite{chen-etal-2023-propsegment}}} \\
\midrule
\textsc{Bert}-Large & $34.97$ & $33.42$ & $34.17$ & $67.42$ & $64.17$ & $65.75$ \\
\textsc{T5}-Large & $55.95$ & $55.05$ & $55.50$ & $78.03$ & $76.74$ & $77.38$ \\

\bottomrule
\end{tabular}
\caption{Sentence segmentation performance of systems used in this paper when evaluated in zero-shot settings on \propsegment. We include the performance of models trained on \propsegment reported by \citet{chen-etal-2023-propsegment} as a reference.}
\label{tab:segment_performance}
\end{table*}

\section{Proposition Segmentation}
\label{app:segment}

In this section, we provide details on the few-shot prompt and distilled T5-large model we use for segmenting sentences into propositions. We provide evaluations the two methods against \propsegment \cite{chen-etal-2023-propsegment}.
\subsection{Prompt for Proposition Segmentation}
\label{app:segment-prompt}

We use the following prompt with \texttt{gpt-3.5-turbo} to generate the initial set of seed training data for segmenting sentences into propositions. We provide one example from \propsegment for in-context learning demonstration. We process $\tilde~23,000$ sentence pairs with the prompt, which generates a total of 44,970 sentences with propositions, after filtering out malformed and empty generations.    
\begin{prompt}{Prompt for sentence $\Rightarrow$ propositions}
\small
Given the following sentence, tell me what claims they are making. Please split the sentence as much as possible, but do not include information not in the sentence.  \\

\textbf{Sentence}: The Andy Warhol Museum in his hometown, Pittsburgh, Pennsylvania, contains an extensive permanent collection of art. \\

\textbf{Claims}: 
\begin{enumerate}
    \item The Andy Warhol Museum is in Pittsburgh.
    \item Andy Warhol's hometown is in Pittsburgh.
    \item Pittsburgh is in Pennsylvania.
    \item The Andy Warhol Museum contains an extensive permanent collection of art.
\end{enumerate}

\textbf{Sentence}: (\textit{input sentence})\\
\textbf{Claims}:
\end{prompt}

\subsection{Training detail of T5 for proposition segmentation}
\label{app:segment-t5}
We finetune a T5-large \cite{2020t5} model on a seed set of training data generated via \texttt{GPT-3.5-turbo}. We use an AdamW optimizer with a constant learning rate of $1e^{-4}$, with a batch size of 128. We train the model for 3 epochs on $8$x Nvidia A6000s, which takes ~2 hours to finish. 

\subsection{Converting propositions from natural language to token masks}
\label{app:segment-convert}
Given a proposition of a sentence in the natural language form, we convert and align it to a subset of tokens from the original sentence with the following steps. We first tokenize and lemmatize each of the tokens in the proposition using NLTK \cite{bird2009natural}. Next, we construct an affinity matrix between the set of lemmatized tokens from the proposition and the sentence. With the matrix, we assign tokens with identical lemmas are assigned a similarity score of 1. To break ties between multiple token matches, we apply a 2D-convolution filter on the affinity matrix, which adds a small score offset for other token matches in a context window of three tokens. With the affinity matrix, we find the optimal alignment between the proposition and sentence tokens with max bipartite matching on the affinity matrix with the Hungarian algorithm \cite{kuhn1955hungarian}.
 
\subsection{Proposition Segmentation Evaluation on \propsegment}
\label{app:propsegment}
To evaluate the quality of propositions extracted via our pipeline, we evaluate the proposition segmentation performance on \propsegment. The results are shown in \autoref{tab:segment_performance}. For details of the Jaccard similarity based evaluation metrics for proposition segmentation, please refer to \citet{chen-etal-2023-propsegment}.

\section{Training and Hyperparameters}
\label{app:hyperparam}
We implement the sub-sentence encoder architecture with \texttt{pytorch} \cite{paszke2019pytorch} and \texttt{pytorch-lightning}~\cite{Falcon_PyTorch_Lightning_2019}.
All of our sub-encoder model variants are trained on 8$\times$ Nvidia A6000 GPUs with 48GB VRAM. 
\paragraph{Distributed Training} Since we adopt in-batch contrastive loss, we scale up the number of negative examples by increasing the batch size with distributed training across GPUs. We distribute training processes across GPU nodes via Distributed Data Parallel (DDP). Specifically, given a minibatch of $N_{gpu} \times M$ sentences, each GPU gets $M$ sentences, which gets forwarded through model parameters on the GPU. Next, we gather and copy all the encoded propositions along with gradients to each of the GPUs, so that each GPU has the full minibatch for loss computation. Each GPU process backpropagates the loss independently on its copy of the model parameters.       

\paragraph{Hyperparameters} For all experiments, we use the temperature parameter $\tau=0.01$ for the supervised contrastive loss, with AdamW optimizer. For Sentence-T5 and GTR variants of the model, we use learning rate of 1$e^{-4}$. For SimCSE, we use learning rate of $5e^{-5}$.  We train the models for 10 epochs, and a linear decay is applied at the end of each epoch, which decreases the learning to 0 after 10 epochs. 
We select the best checkpoint based on validation loss after each epoch.  

\section{Evaluation Setup}
\label{app:evaluation}
\subsection{Representing Atomic Propostion with Sentence Encoder}
With the atomic fact retrieval evaluation on \propsegment, since the ground truth query and target propositions are both represented in the format of token masks, we experiment with a few different strategies of formatting the input for sentence encoders. Specifically, with respect to the input sentence and the token masks denoting the proposition, here are the different strategies in consideration. 
\begin{enumerate}
    \item \textbf{Mask pooling only.} Encoder has full attention, apply proposition mask during pooling. Note that this is the same method we use for the sub-sentence encoder.
    \item \textbf{Full mask.} Apply proposition mask as attention mask during both encoding and pooling.
    \item \textbf{Token subset only.} Take the subset of tokens and discard the rest. Feed only the subset of tokens as a sequence to the encoder and pooling layer. 
\end{enumerate}
When tested on a small validation set for the atomic fact retrieval task, we generally observe that \textbf{mask pooling only} yields the best result across most models, except for the two compact models, i.e. MiniLM-L6-v2 and DistilRoberta. On the two compact models, we see \textbf{full mask} outperforming the \textbf{mask pooling only} strategy by a small margin. We observe that with the third strategy \textbf{token subset only}, the validation performance trails behind the other two across all models.  

\end{document}